\documentclass[sigconf]{acmart}

\usepackage{algorithm,algorithmic}

\AtBeginDocument{%
  \providecommand\BibTeX{{%
    \normalfont B\kern-0.5em{\scshape i\kern-0.25em b}\kern-0.8em\TeX}}}

 \setcopyright{iw3c2w3}

\acmSubmissionID{ECNLP06s}

\begin{document}

\copyrightyear{2020}
\acmYear{2020} 
\acmConference[WWW '20 Companion]{Companion Proceedings of the Web Conference 2020}{April 20--24, 2020}{Taipei, Taiwan}
\acmBooktitle{Companion Proceedings of the Web Conference 2020 (WWW '20 Companion), April 20--24, 2020, Taipei, Taiwan}
\acmPrice{}
\acmDOI{10.1145/3366424.3386196}
\acmISBN{978-1-4503-7024-0/20/04}

\title{Interpretable Methods for Identifying Product Variants}

\author{Rebecca West}
\affiliation{%
   \institution{The Home Depot}
   \city{Atlanta}
   \state{GA}
   \country{USA}}
\email{rebecca_west@homedepot.com}

\author{Khalifeh Al Jadda}
\affiliation{%
   \institution{The Home Depot}
   \city{Atlanta}
   \state{GA}
   \country{USA}}
\email{khalifeh_al_jadda@homedepot.com}

\author{Unaiza Ahsan}
\affiliation{%
   \institution{The Home Depot}
   \city{Atlanta}
   \state{GA}
   \country{USA}}
\email{unaiza_ahsan@homedepot.com}

\author{Huiming Qu}
\affiliation{%
   \institution{The Home Depot}
   \city{Atlanta}
   \state{GA}
   \country{USA}}
\email{huiming_qu@homedepot.com}

\author{Xiquan Cui}
\affiliation{%
   \institution{The Home Depot}
   \city{Atlanta}
   \state{GA}
   \country{USA}}
\email{xiquan_cui@homedepot.com}


\begin{abstract}
  For e-commerce companies with large product selections, the organization and grouping of products in meaningful ways is important for creating great customer shopping experiences and cultivating an authoritative brand image. One important way of grouping products is to identify a family of product variants, where the variants are mostly the same with slight and yet distinct differences (e.g. color or pack size). In this paper, we introduce a novel approach to identifying product variants. It combines both constrained clustering and tailored NLP techniques (e.g. extraction of product family name from unstructured product title and identification of products with similar model numbers) to achieve superior performance compared with an existing baseline using a vanilla classification approach. In addition, we design the algorithm to meet certain business criteria, including meeting high accuracy requirements on a wide range of categories (e.g. appliances, decor, tools, and building materials, etc.) as well as prioritizing the interpretability of the model to make it accessible and understandable to all business partners.
\end{abstract}





\maketitle

\section{Introduction}
\begin{figure}[h]
  \centering
  \includegraphics[width=\linewidth]{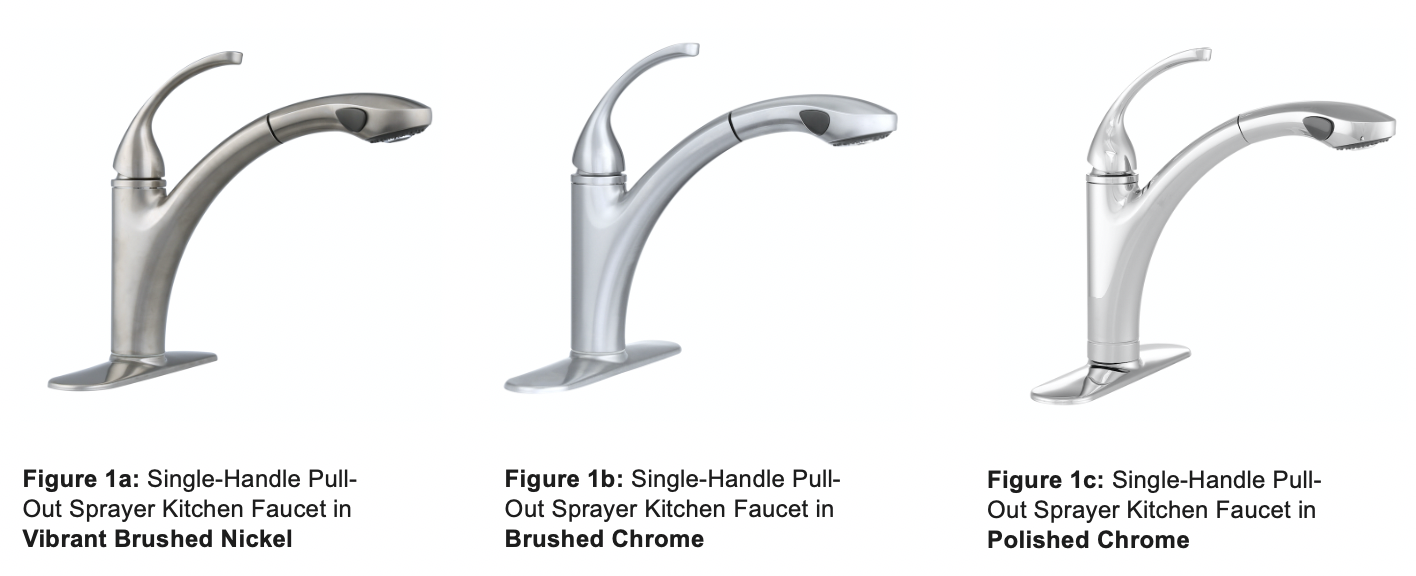}
  \caption{The same faucet is sold in various finishes.}
  \Description{Product variants for a faucet}
  \label{variant_example}
\end{figure}

The organization and grouping of products is important for e-commerce and retail companies to streamline the shopping experience and showcase product authority. Identifying product variants is one of the important tasks within the domain of product grouping and recommendation. Product variants are groups of products that share many similarities except for a few differences. For example, the same faucet may be sold in a number of different finishes, as shown in Figure~\ref{variant_example}.

Product variants are particularly useful to customers for a number of reasons.
\begin{itemize}
    \item They help users find alternative products easily and without the hassle of searching or browsing through the massive product catalog.
    \item They save consumers time in comparing alternative products by ensuring that key product features remain the same across product variants.
    \item They simplify search results and product lists by grouping similar products together.
\end{itemize}

\begin{figure}[h]
  \centering
  \includegraphics[width=\linewidth]{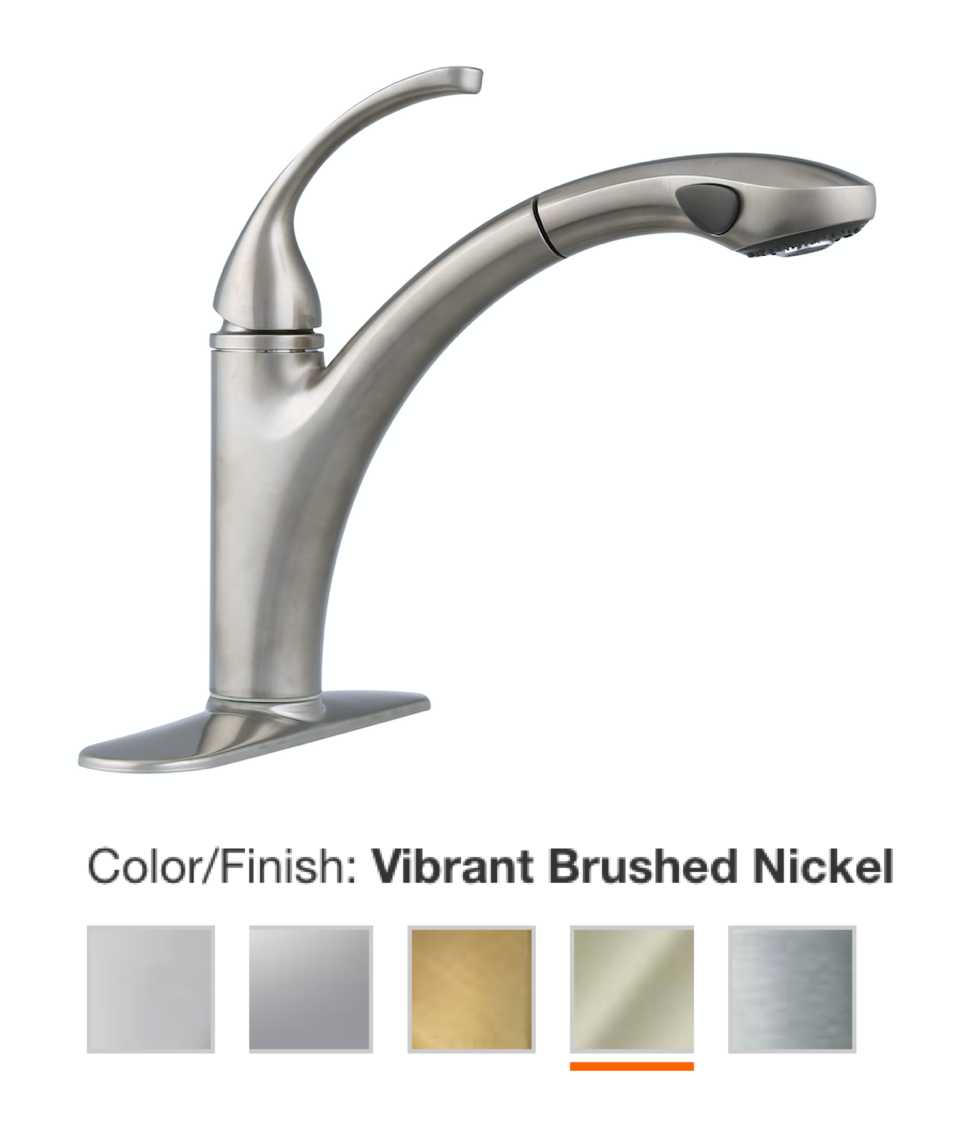}
  \caption{Product variant options are prominently displayed on the website.}
  \Description{Product variant information prominently displayed on website}
  \label{website_swatch}
\end{figure}

On e-commerce websites, product variant information is often prominently displayed on product detail and product listing pages, and receives extraordinarily high user views, clicks, and purchases.  Figure~\ref{website_swatch} shows how product variant options may be shown to the user. 

Due to the importance of product variant identification, the execution requirements for the algorithm are very high. In addition to high performance, the algorithm must have a high degree of interpretability and transparency to receive support from business stakeholders and enable collaboration with various domain experts.

\subsection{Algorithm Requirements}
\subsubsection{Precision}
Product variant information is highly visible to end customers, and mistakes can significantly impact customer experience and site performance. As a result, there is a very high requirement for the precision of product variant identification.

\subsubsection{Recall}
High recall is also highly desired. Showing the complete product offerings not only helps customers find the right products quicker, but also projects the right image of product authority and increases customer loyalty. 

\subsubsection{Leverage Domain Knowledge}

When working with a wide variety of product categories, it is important to define product variants in different ways. For example, while faucets may vary by color, it does not make sense to define lumber variants based on color. For many types of products, there are existing business rules for how variants in each product category should be defined. In order to achieve the highest level of customer satisfaction, our algorithm must respect this domain knowledge.

\subsubsection{Interpretability}
Identifying product variants is a high priority task that involves many business stakeholders. In order to effectively collaborate with them, we need a business friendly justification for \textit{why} certain products were grouped together. When business users have a clear understanding of why the algorithm is grouping certain products, they are more likely to adopt the algorithm on their set of products or contribute their subject matter expertise towards improving the algorithm.

Based on the above requirements, we frame the problem of identifying product variants into a constrained clustering problem, where we incorporate business rules and human knowledge into the algorithm as constraints. In order to improve recall, we use Natural Language Processing (NLP) techniques to perform sensible fuzzy matching on the constraints to overcome the noise in the textual data. First, we extract product family name from unstructured product title. Then, we measure the similarity between model numbers of two product variant candidates using edit distance. Beside several business constraints, there are cluster level constraints that the product variants must have the same family name and the model number difference must be within a certain range. We will discuss the details in the Section 3.

\section{Related Work}

In this section, we discuss three types of strategies that are relevant to our application, including product deduplication, classification, and constrained clustering.

\subsection{Product Deduplication}

Product duplication broadly involves detecting duplicates or similar products in databases. Several methods address this problem as it is extremely important in retail applications. \cite{alenazi2016record} provides a comprehensive overview of database duplicates detection. Product similarity can be computed in various ways. 
\paragraph{Character-based methods:} These methods match strings character by character. Popular methods that do character-wise matching include edit distance or Levenshtein Distance \cite{levenshtein1966binary} and Hamming Distance. 
\paragraph{Token-based methods:} Character-based methods may not work well if the order of words get changed (but they mean the same). Hence token-based methods are useful where raw strings are not matched but are converted to tokens first (by splitting them using a delimiter). Jaccard index is a popular token-based similarity metric. 

Besides databases, product duplicate detection is addressed for Web data obtained from e-commerce websites. The key difference between finding similar products from a relational database and from the Web is that Web data is very noisy whereas databases are structured. \cite{van2015multi} propose a hierarchical clustering method to cluster products from multiple websites and apply it to detect similar products in a dataset of TVs. Their approach builds upon the Title Model Words Method (TMWM) \cite{vandic2012faceted} which extracts entities from product titles and matches them using cosine similarity. Then additional rules are applied to detect product duplicates. Another method that their approach uses is the Hybrid Similarity Method (HSM) \cite{de2013hybrid} which adds product attributes to compute similarities (beyond just product titles). 

Another relevant line of work \cite{strauss2019applying} uses the term product resolution and heuristic methods (or rules) to extract structure from unstructured data and find product duplicates. \cite{hassanian2019pruning} is also a rule-based filtering approach that does pairwise comparison between punctuation-delimited textual chunks and determines if the chunks are similar or not. Pairwise computations save time especially if the dataset is large. Another method that builds upon this idea is \cite{hartveld2018lsh} where the authors propose to use Locality Sensitive Hashing (LSH) \cite{har2012approximate} to detect candidates for pairwise comparison and compare model words from product titles (proposed by \cite{van2016duplicate}) as well as descriptions. 

Despite of the large body of work on product deduplication, these methods do not directly address our application. Rather than identifying the most similar products, our task is to group items together conditional on variable attributes defined by domain experts for each specific category. This uncompromisable condition makes the task more challenging and interesting. 

\subsection{Classification}

A recent blog post \cite{Maunu01} proposed a classification based approach to find product variants. They predict whether a pair of products belong to the same product variant family based on a limited set of generic product attributes. These attributes are the same across all categories and include dimensions, price, weight, manufacturer, supplier, and model number. For each of these attributes, a distance metric is computed between each pair of items. These distance metrics are used as features to train a random forest model, which gives a probability of product similarity.

One drawback of this classification method is that it uses the same standard attributes for all categories. When the product catalog is very diverse, we found that classification algorithms using these standard attributes do not perform well. In particular, the attributes may be unavailable or not relevant to certain types of products.

\begin{figure}[!ht]
    \centering
    \includegraphics[width=\linewidth]{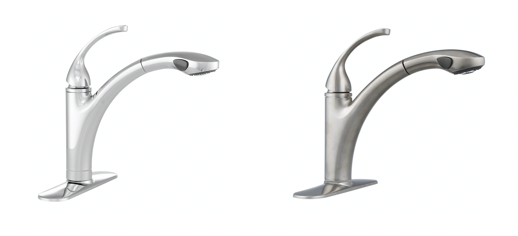}
    \qquad
      \begin{tabular}{cl}
    \toprule
    Feature Name&Feature Value\\
    \midrule
    Weight Difference & NULL\\
    Length Difference & NULL\\
    Width Difference & NULL\\
    Depth Difference & NULL\\
    Price Difference & 24.2\% \\
    Model Number Difference & 2\\
  \bottomrule
\end{tabular}
    \captionlistentry[table]{Features for faucet pair}
    \caption{Vibrant brushed nickel and polished chrome faucets are compared using standard classification features. In this baseline model, these items are not correctly identified as product variants.}
    \label{tab:wayfair}
  \end{figure}

Many categories may not have standard attributes like Product Width or Product Length defined. These missing values degrade the performance of the model. We used standard techniques (e.g. average or dominant values of similar products) to fill in missing values. In addition, some products, such as windows, are designed to vary on these standard dimension attributes. Therefore, it often does not make sense to group based on similarity within these attributes. Figure~\ref{tab:wayfair} shows two faucets that should be classified as within the same product variant family. The distance features used in the model are shown in the table. In this example, many of the standard product dimensions have missing values and the faucets are not correctly identified as being product variants.

In contrast, our algorithm only uses two basic product attributes, product title and model number. Because they are required to make a product sellable, we rarely suffer from the missing value issue. In addition, our algorithm considers variable attributes defined by domain experts for each category in the process of extracting the product family name (will be discussed in Section 3.2). Therefore, our algorithm generated better results that satisfy the category-specific conditions. 

\subsection{Constrained Clustering}
Constrained clustering \cite{tung2001constraint} is a type of clustering algorithm that considers the user's background knowledge or constraints in the processing of grouping objects into clusters. This special property makes it more desirable than its unconstrained counterpart in many applications. The standard definition of constrained clustering is similar to typical unconstrained clustering models with the addition of a set of constraints imposed on each cluster, as shown below:

\theoremstyle{definition}
\begin{definition}{\textbf{Constrained Clustering:}}
Given a data set $D$ with $n$ objects, a distance function $d : D \times D \rightarrow \mathbb{R}$, a positive integer $k$, and \textit{a set of constraints} $\mathcal{C}$, find a \textit{k-clustering} $(Cl_1, ..., Cl_k)$ such that 
$$DISP = (\sum_{i = 1}^{k} disp(Cl_i, rep_i))$$
is minimized, and \textit{each cluster} $Cl_i$ \textit{satisfies the constraints} $\mathcal{C}$, denoted as $Cl_i \models \mathcal{C}$.

$disp(Cl_i, rep_i) = \sum_{O \in Cl_i} d(O, rep_i)$ represents the dispersion of the cluster $Cl_i$ or the total distance between every member of the cluster and a representative element $rep_i \in Cl_i$ of the cluster.
\end{definition}

A type of constrained clustering is based on SQL aggregate constraints, defined as: 

\theoremstyle{definition}
\begin{definition}{\textbf{SQL Aggregate Constraints:}}
Consider the aggregate functions $agg \in \{max(), min(), avg(), sum()\}$. Let $\theta$ be a comparator function, i.e., $\theta \in \{<, \leq, \neq, =, \geq, >\}$ and $c$ represent a numeric constant. Let $O_i \in D$ be associated with attributes $\{A_j\}$ where $O_i[A_j]$ represents the value of attribute $A_j$ for object $O_i$. Given the cluster $Cl$, a SQL aggregate constraint on $Cl$ is a constraint in one of the following forms:
\renewcommand{\labelenumi}{\roman{enumi}}
\begin{enumerate}
    \item $agg(\{O_i[A_j]|O_i \in Cl\}) \theta c$;
    \item $count(Cl) \theta c$
\end{enumerate}
\end{definition}

In a sense, this paper is about how to use NLP approaches to create proper SQL aggregate constraints (i.e. satisfying business rules, sharing the same product family name, and model numbers within a certain edit distance) to perform constrained clustering. Because this approach respects the business requirements and uses easily interpretable features, it achieves satisfactory results and accelerates its adoption in the business. However, it is worth noting that our algorithm is different from the traditional constrained clustering approach. From the definition, constrained clustering requires a pre-defined parameter k for the number of clusters. In contrast, our algorithm does not limit on the number of product variant groups, and can allow the formation of product variant groups to adapt to the properties of each category. This is the most sensible approach given the business use case.

\section{Approach}

In this section, we describe our approach to identifying product variants using the framework of constrained clustering. A summary of this approach can be found in Algorithm~\ref{algo:recommendation}. We describe in detail the three main constraints that we impose on our constrained clustering algorithm.

\begin{itemize}
    \item \textbf{Business rule constraints} - We require an exact match on several business mandated attributes. For each product variant group or cluster $Cl$, 
    $$O_i[A_j] = O_k[A_j]$$
    for all $O_i, O_k \in Cl$ and $A_j \in \{brand, \allowbreak category\}$.
    \item \textbf{Same product family name constraint} - For each cluster $Cl$, 
    $$O_i[A_j] = O_k[A_j]$$
    for all $O_i, O_k \in Cl$, where $A_j$ represents the attribute of product family name. In the next section, we will talk about the NLP process to extract the attribute of product family name from the product title. This is conditional on the variant attributes set by domain experts.
    \item \textbf{Model number constraint} - The model number for each product must be similar to the model number of another product in the cluster. For each cluster $Cl$, 
    $$\min_{O_k \in Cl}{d(O_i[A_j], O_k[A_j]}) \leq c$$
    for all $O_i \in Cl$ where $A_j$ represents the model number attribute and $d$ represents the Levenshtein distance function.
\end{itemize}

These constraints are further described in the sections below and a summary of our algorithm is given in Algorithm~\ref{algo:recommendation}.

\subsection{Rule Based Constraints}

Product variants must be grouped from within the same brand and category due to business constraints.

\subsection{Extraction of Family Name from Product Title Conditional on Category-specific Variant Attributes}

Our algorithm extracts a product family name for each product. The family name consists of a sequence of important tokens from the product title that should be found in all products within a product variant family. This family name allows us to easily explain to business partners how the products were grouped. We use the following steps to extract the product family name from the product title.

\begin{itemize}
    \item Standardize terms with a synonyms dictionary.
    \item Remove punctuation and other non-standard characters.
    \item Remove numbers and units of measurement as these are often noisy and inconsistent.
    \item Remove brand information as this is inconsistently available in the title.
    \item Remove noisy tokens from a manually created blacklist, including tokens from noisy attributes.
    \item Remove tokens that are already included in the category data.
    \item Remove tokens related to the variant attributes that are defined by domain experts for each category. For example, the color of the faucet should be removed from the title.
\end{itemize}

\begin{figure}[h]
  \centering
  \includegraphics[width=\linewidth]{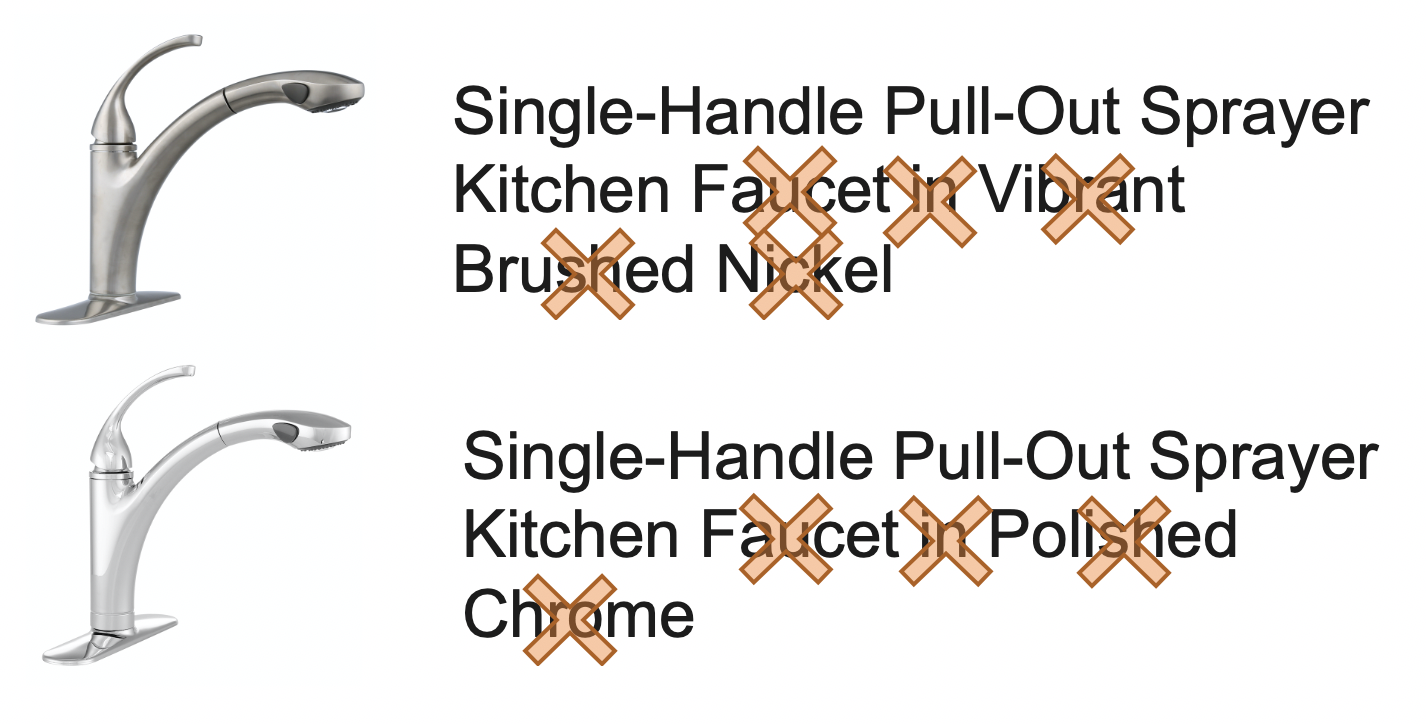}
  \caption{Product family name is extracted by multiple NLP steps, e.g. normalization and removing tokens that are noisy or are from variable attributes defined by domain experts.}
  \Description{Product title cleaning example}
  \label{title_cleaning}
\end{figure}

Figure~\ref{title_cleaning} shows an example of how a product family name is extracted from a unstructured product title. If product candidates have the same sequence of tokens after the cleaning process, these products may belong to the same product variant group.

\subsection{Model Number}

\begin{figure}[h]
  \centering
  \includegraphics[width=\linewidth]{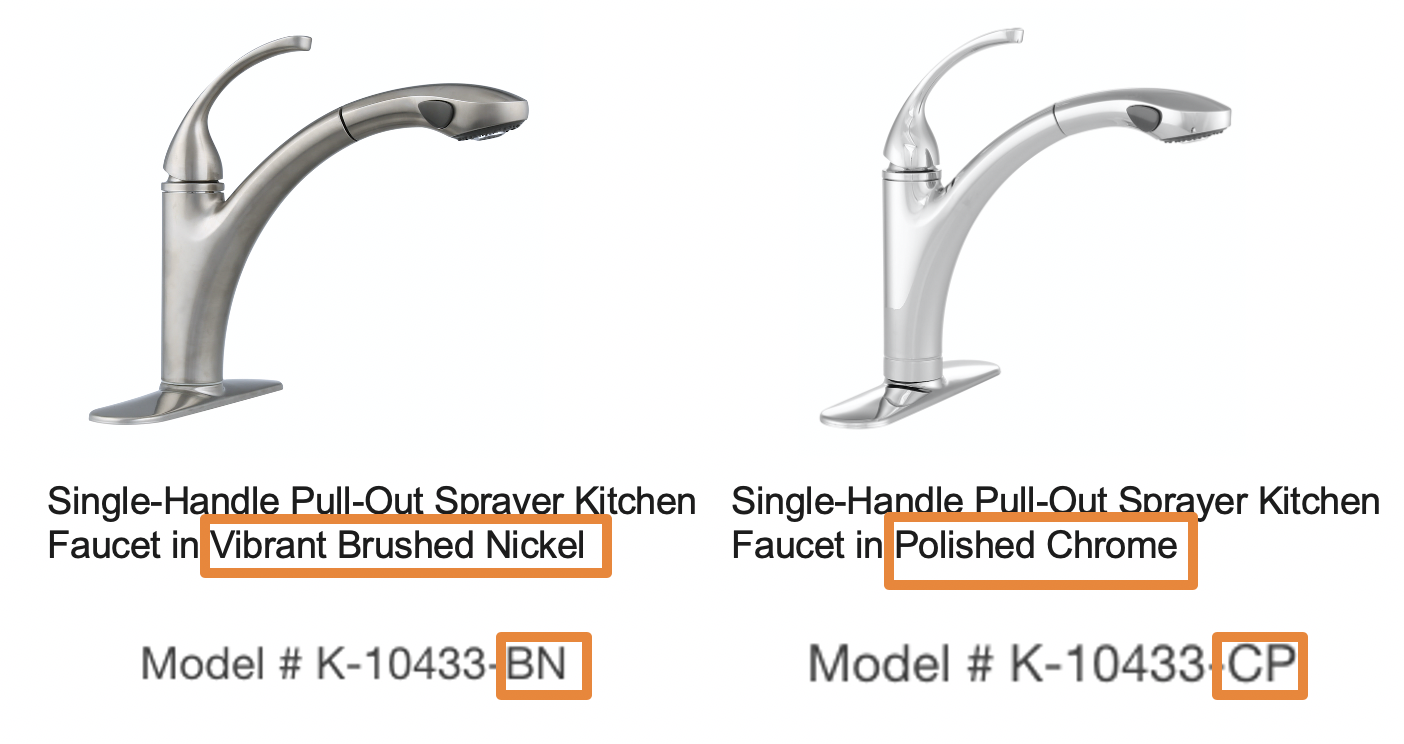}
  \caption{Product variants are often assigned similar model numbers by the manufacturer. In this example, BN stands for Brushed Nickel and CP stands for Polished Chrome.}
  \Description{Model number similarities.}
  \label{model_no_ex}
\end{figure}

Another strong signal that products belong to the same product variant family is the model number. Product variants are often assigned similar model numbers by the manufacturer. The core part of the model number is often constant for all product variants, while a few characters may be changed or added to represent the available attributes. In Figure~\ref{model_no_ex}, we see that the faucet color is represented by the two final letters in the model number. All other characters in the model number remain constant.

We use the standard Levenshtein or edit distance to measure how similar two model numbers are. The following operations are included with equal weights:
 \begin{itemize}
     \item Insertion
     \item Deletion
     \item Substitution
 \end{itemize}
 
\begin{figure}[h]
  \centering
  \includegraphics[width=\linewidth]{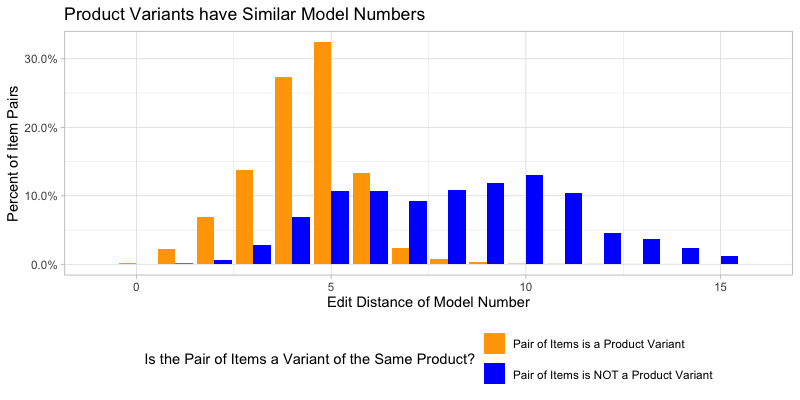}
  \caption{The edit distance between model numbers is closer for items that are product variants than for two different items.}
  \Description{Model number edit distance.}
  \label{model_no_dist}
\end{figure}

This distance measure helps us determine if a small change or addition was made between product variants. In general, product variants tend to have much smaller edit distances than pairs of products that are from a similar category but are not product variants. Figure~\ref{model_no_dist} shows the distribution of edit distances for product variants compared to different products from within the same category. Product variants tend to have lower model number edit distances.

How model numbers are created tends to vary by category and manufacturer. In particular, different categories may have different standards in how similar model numbers are. We used a grid search to determine the optimal model number distance threshold for each of our major categories.

\begin{algorithm}
    \caption{Finding Product Variants}
    {\fontsize{10}{10}\selectfont
    \begin{algorithmic}[1]
        \renewcommand{\algorithmicrequire}{\textbf{Input:}}
        \renewcommand{\algorithmicensure}{\textbf{Output:}}
        \REQUIRE Products $p_i, p_j  \in \mathcal{P}$,
        \newline $\mathcal{P}$ the set of $n$ products,
        \newline $p_i[brand]$ the brand of product $p_i$
        \newline $p_i[category]$ the category for product $p_i$
        \newline $h(p_{i}[title])$ the product family name for $p_{i}$, extracted from the product title
        \newline $d(p_i[model\ number], p_j[model\ number])$ the Levenstein distance between the model numbers for products $p_i$ and $p_j$
        \newline $c_i$ the cutoff for how far apart model numbers may be for all products of the same product type as $p_i$
        \ENSURE A set of product variants: $$\{p_i : p_i \text{are all variants of the same product}\}$$
        \STATE Check that $p_i[brand] = p_j[brand]$
        \STATE Check that $p_i[category] = p_j[category]$
        \STATE Check that $h(p_{i}[title]) = h(p_{j}[title])$
        \STATE Check that $d(p_i[model\ number], p_j[model\ number]) \leq c_i$
        \RETURN Are $p_i$ and $p_j$ product variants?
    \end{algorithmic}
    }
    \label{algo:recommendation}
\end{algorithm}

\section{Experiments}

In addition to the required business constraints, brand and category, we test the following two combination of constraints:

\begin{itemize}
    \item \textbf{Model Number Only} - Items are grouped together as product variants if the edit distance of their model numbers is sufficiently small.
    \item \textbf{Extracted Product Family Name and Model Number} - Important tokens are extracted from each product title. Items that have the same product title tokens and also have a sufficiently small edit distance for the model number are grouped together as product variants.
\end{itemize}

We compare the performance of constrained clustering under these constraints to a baseline of vanilla classification.

We test these methods using a manually curated dataset of product variants owned by a large retailer. We include a wide variety of product categories such as window treatments, faucets, PVC pipes, tools, hardwood flooring, and many other products. This dataset consists of about 1 million items.

Precision is measured by comparing items grouped by the algorithm to items with known product variant information. True positives are when our algorithm groups together items that are known to be variants of the same product. False positives occur when our algorithm groups together items that are known to be variants of different products. Products with no known variant information are ignored in this validation process. Recall is measured by looking at items with known product variant groupings, and determining how many of these items are correctly identified. Model performance is shown in Table~\ref{tab:f1}.

The model number constraint is not sufficient to correctly identify product variants. This method performs poorly compared to the classification baseline. However, when the product family name constraint is added, we see a significant increase in recall resulting in a much higher F1 score than the baseline.

\begin{table*}
  \caption{Model Performance}
  \label{tab:f1}
  \begin{tabular}{ccccl}
    \toprule
    Model&Precision&Recall&F1 Score&Percent of Highly Accurate Categories\\
    \midrule
    Classification Baseline & \textbf{64\%} & 33\% & 0.44 & 35.7\% \\
    Model Number Only & 32\% & 6\% & 0.10 & 23.3\% \\
    Cleaned Title and Model Number & 62\% & \textbf{92\%} & \textbf{0.74} & \textbf{51.3\%} \\
  \bottomrule
\end{tabular}
\end{table*}

\begin{figure}[h]
  \centering
  \includegraphics[width=\linewidth]{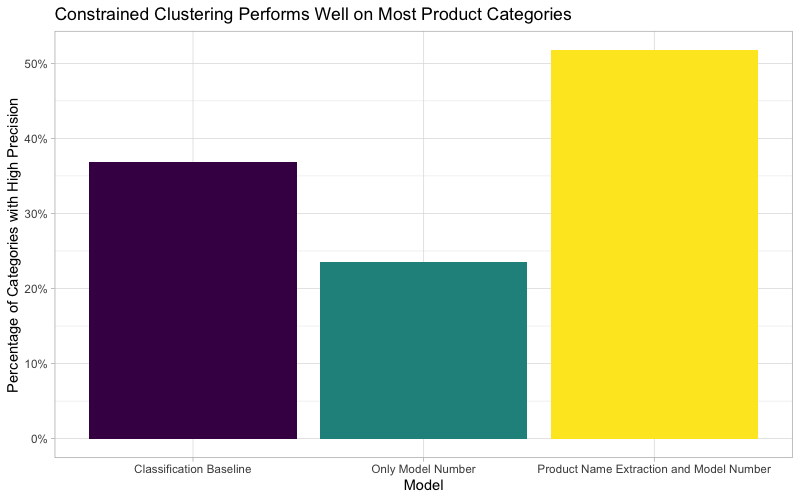}
  \caption{Percent of categories which reached the high precision threshold for each algorithm.}
  \Description{Category-specific precision}
  \label{tax_dist}
\end{figure}

In addition to overall precision and recall, we review the performance of our model across different categories. Algorithm results are released into production on a category by category basis. Due to business requirements, we can only put the model into production for categories that have very high precision (>=90\%). Using both the same product family name and similar model number constraints resulted in a much higher number of categories reaching the precision threshold than the classification model, as shown in Figure~\ref{tax_dist}. This results in the largest number of groupings being implemented on the site.

\section{Conclusion}

While there are many techniques for identifying similar or duplicate products, not all of these methods are appropriate for grouping product variants within a diverse catalog. A constrained clustering approach gives us the flexibility to group items from a wide variety of product types while still maintaining high accuracy. Constrained clustering also provides an easy to understand explanation for why products are grouped together. The most effective constraints are based on explicitly incorporating business rules, extracting the product family name from the product title, and making use of patterns encoded in the model number by the manufacturer. Applying these constraints to our product catalog gives us high quality product variant groupings.

\bibliographystyle{ACM-Reference-Format}


\end{document}